\documentclass[journal]{IEEEtran}
\usepackage[T1]{fontenc}
\usepackage{graphicx}
\usepackage{times}
\usepackage{helvet}
\usepackage{courier}
\usepackage{amsmath}
\usepackage{algorithm}
\usepackage{algorithmic}
\usepackage{csquotes} 
\usepackage{color}
\usepackage{paralist}
\usepackage{amssymb}
\usepackage{indentfirst}
\usepackage{subfigure}
\usepackage{float}
\usepackage{multirow}
\usepackage{cite}
\usepackage{mathrsfs}

\usepackage{mathrsfs} 
\usepackage{amsfonts}
\usepackage{todonotes}
\usepackage{pgfplots} 
\usepackage{epstopdf}
\usepackage{epsfig}
\pgfplotsset{compat=newest}
\usepackage{color}
\definecolor{forestgreen}{RGB}{0,139,69}

\usepackage{xcolor}
\definecolor{citecolor}{HTML}{0071bc}
\usepackage[colorlinks, linkcolor=red,  anchorcolor=blue, citecolor=citecolor]{hyperref} 

\usepackage{xcolor}
\definecolor{SeaGreen4}{RGB}{0,205,102} 
\definecolor{SlateBlue}{RGB}{106,90,205} 
\definecolor{DarkRed}{RGB}{178,34,34} 
	
\usepackage[switch]{lineno}

\usepackage{textcomp,booktabs}
\usepackage{amssymb}
\usepackage{pifont}
\newcommand{\cmark}{\ding{51}}%

\usepackage{makecell}

\usepackage{colortbl}
\definecolor{mygray}{gray}{.9}
\definecolor{mypink}{rgb}{.99,.91,.95}
\definecolor{mycyan}{cmyk}{.3,0,0,0}

\begin{document}

\title{VELoRA: A Low-Rank Adaptation Approach for Efficient RGB-Event based Recognition}    

\author{Lan Chen, Haoxiang Yang, Pengpeng Shao, Haoyu Song, Xiao Wang*, \emph{Member, IEEE}, \\ 
        Zhicheng Zhao, Yaowei Wang, \emph{Member, IEEE}, Yonghong Tian, \emph{Fellow, IEEE}  
\thanks{$\bullet$ Lan Chen is with the School of Electronic and Information Engineering, Anhui University, Hefei 230601, China. (email: chenlan@ahu.edu.cn)}  
\thanks{$\bullet$ Xiao Wang, Haoxiang Yang, and Haoyu Song are with the School of Computer Science and Technology, Anhui University, Hefei, China. (email: xiaowang@ahu.edu.cn, 15357237278@163.com, songhaoyuu@gmail.com	)}  
\thanks{$\bullet$ Pengpeng Shao is with Tsinghua University, Beijing, China. (email: ppshao@tsinghua.edu.cn)} 
\thanks{$\bullet$ Zhicheng Zhao is with the School of Artificial Intelligence, Anhui University, Hefei, China. (email: zhaozhicheng@ahu.edu.cn)} 
\thanks{$\bullet$ Yaowei Wang is with Peng Cheng Laboratory, Shenzhen, China, and Harbin Institute of Technology, Shenzhen, China. (email: wangyw@pcl.ac.cn)} 
\thanks{$\bullet$ Yonghong Tian is with Peng Cheng Laboratory, Shenzhen, China, and National Engineering Laboratory for Video Technology, School of Electronics Engineering and Computer Science, Peking University, Beijing, China. (email: yhtian@pku.edu.cn)} 
\thanks{* Corresponding author: Xiao Wang}   
}

\markboth{ IEEE Transactions on ***, 2024 } 
{Shell \MakeLowercase{\textit{et al.}}: Bare Demo of IEEEtran.cls for IEEE Journals}

\maketitle

\begin{abstract}
Pattern recognition leveraging both RGB and Event cameras can significantly enhance performance by deploying deep neural networks that utilize a fine-tuning strategy. Inspired by the successful application of large models, the introduction of such large models can also be considered to further enhance the performance of multi-modal tasks. However, fully fine-tuning these models leads to inefficiency and lightweight fine-tuning methods such as LoRA and Adapter have been proposed to achieve a better balance between efficiency and performance. To our knowledge, there is currently no work that has conducted parameter-efficient fine-tuning (PEFT) for RGB-Event recognition based on pre-trained foundation models. 
To address this issue, this paper proposes a novel PEFT strategy to adapt the pre-trained foundation vision models for the RGB-Event-based classification. Specifically, given the RGB frames and event streams, we extract the RGB and event features based on the vision foundation model ViT with a modality-specific LoRA tuning strategy. The frame difference of the dual modalities is also considered to capture the motion cues via the frame difference backbone network. These features are concatenated and fed into high-level Transformer layers for efficient multi-modal feature learning via modality-shared LoRA tuning. Finally, we concatenate these features and feed them into a classification head to achieve efficient fine-tuning. 
The source code and pre-trained models will be released on \url{https://github.com/Event-AHU/VELoRA}. 
\end{abstract}

\begin{IEEEkeywords}
Efficient Fine-tuning; Pattern Recognition; Event Camera; Visible-Event Fusion; Multi-modal Fusion 
\end{IEEEkeywords}

\IEEEpeerreviewmaketitle

\section{Introduction}

\IEEEPARstart{P}{attern} recognition targets learning and finding the common patterns from the given data directly and helps the downstream tasks, such as object detection~\cite{10098596, xie2024event}, \cite{ren2024dino} and tracking~\cite{7780603, ramesh2020etld}, \cite{ren2016faster}, \cite{5255236}, segmentation~\cite{karimi2021convolution, zhu2024continuous}, \cite{cao2022swin}, etc. This research direction achieves significant improvements with the help of deep neural networks. However, the mainstream algorithms developed based on RGB cameras perform poorly in extremely challenging scenarios (e.g., low illumination and fast motion) due to the limited dynamic range and frame rate. Therefore, pattern recognition is still far from being solved in these cases.

To address the issues mentioned above, some researchers exploit bio-inspired event cameras (also termed Dynamic Vision Sensors, DVS) to assist RGB cameras for more accurate pattern recognition. From the perspective of the imaging principle, event cameras only emit a pulse/spike when the variation of light intensity exceeds a threshold. Note that, all pixel points are emitted asynchronously, meaning that whether each pixel point records information is independent and does not interfere with each other. Traditional frame cameras, on the other hand, record all pixel points in a scene synchronously. Event cameras have a higher dynamic range and higher temporal resolution, allowing them to work well in low-light or high-exposure scenes, with almost no motion blur. In addition, event cameras also have the advantage of low power consumption, which is beneficial for deployment on devices with limited energy consumption. The sparse spatial resolution allows for better handling of privacy-sensitive issues, such as human-centric visual tasks (pedestrian attribute recognition, person re-identification).

\begin{figure}
\centering
\includegraphics[width=0.9\linewidth]{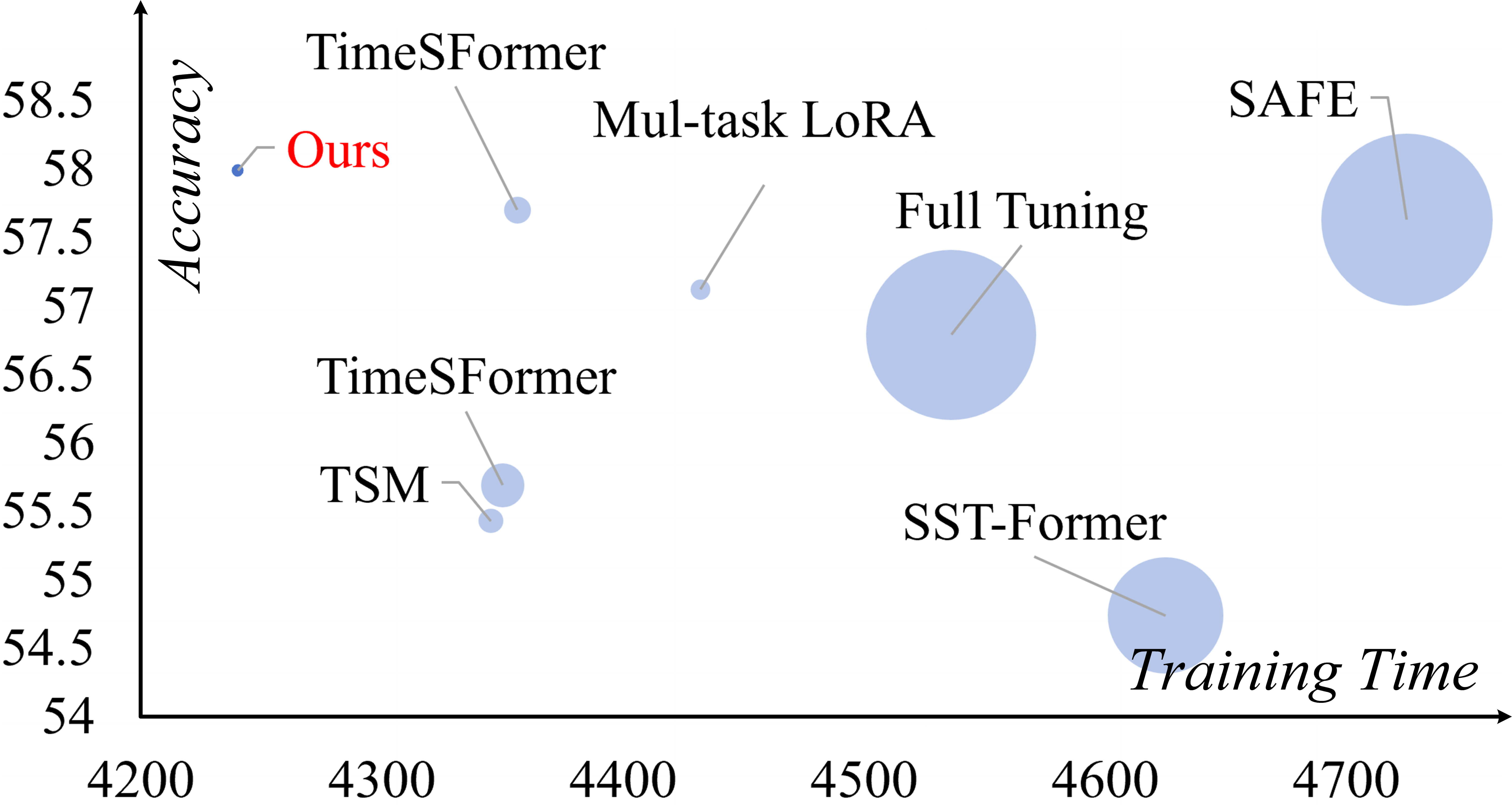}
\caption{Comparison between existing event-based classification models and our newly proposed VELoRA. 
The horizontal axis represents \textit{training time}, the vertical axis represents the \textit{accuracy} of the model, and the size of the bubble indicates the \textit{amount of parameters} that need to be adjusted.} 
\label{fig:firstIMG}
\end{figure}

More in detail, 
Li et al. propose the SAFE~\cite{li2023semantic} which exploits the fusion of RGB frames, event streams, and category names for high-performance RGB-Event based pattern recognition. 
Wang et al. propose a spatial-temporal feature learning framework based on Transformer, i.e., ESTF~\cite{wang2024hardvs}, for event-based human activity recognition. 
Chen et al.~\cite{chen2023ECSNet} propose a 2D-1T event cloud sequence which is a compact event representation for spatio-temporal information encoding. 
Deng et al.~\cite{deng2022MVFNet} project the event stream into multi-view 2D maps using a multi-view attention-aware network for spatio-temporal feature learning. 
Although these works achieve good performance on the event-based classification, however, we believe these models are limited by the following issues. 
\textbf{Firstly}, these models use weights obtained from training on large-scale classification datasets or image-text datasets as their starting parameters and employ full fine-tuning to achieve better results. Obviously, this fine-tuning strategy limits the training efficiency of the model. Although researchers have proposed various lightweight fine-tuning methods such as LoRA~\cite{hu2021lora}, Side Net~\cite{zhang2020side}, \cite{chen2022vision}, and Adapter~\cite{chen2022adaptformer}, \cite{pan2022st} to better adapt vision foundation models/LLMs/multi-modal large models to pattern recognition tasks, these methods cannot be directly applied to the multi-modal fusion framework. 
\textbf{Secondly}, some variants of parameter-efficient fine-tuning strategies, such as MoE-LoRA~\cite{feng2024mixture} and MTLoRA~\cite{agiza2024mtlora}, have achieved good results in multi-task/multi-expert fine-tuning. However, our experimental results indicate that these models are less effective in multi-modal fusion.
Thus, it is natural to raise the following question: \emph{how can we design an efficient fine-tuning strategy specifically tailored for RGB-Event-based multi-modal fusion to better adapt visual foundation models?}

\begin{figure*}
\centering
\includegraphics[width=0.85\textwidth]{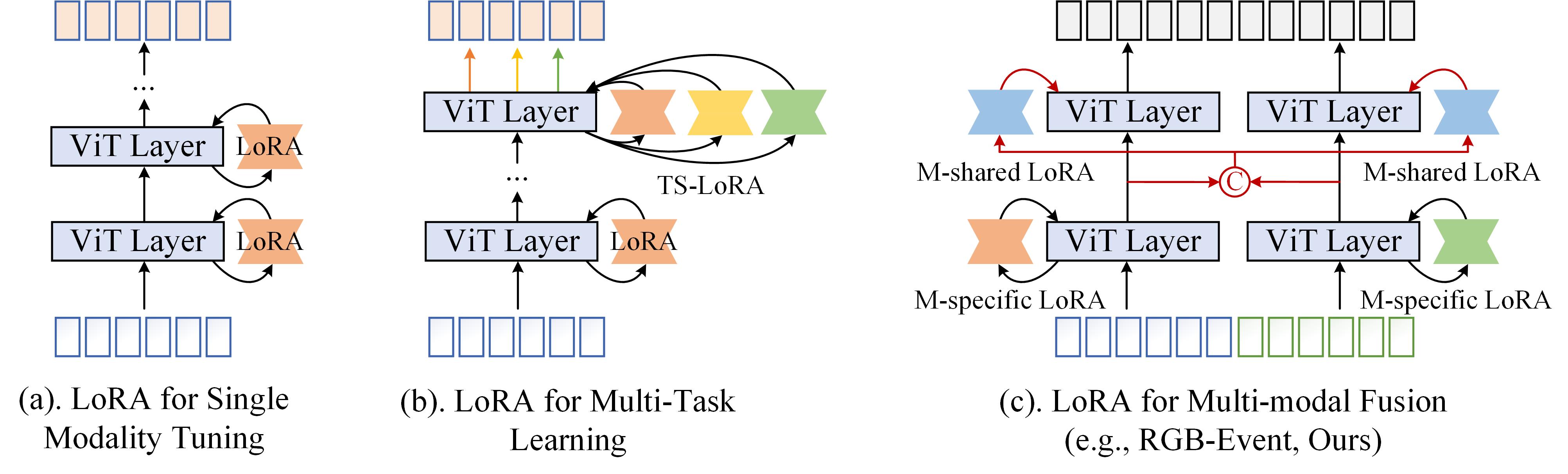}
\caption{Comparison between existing LoRA strategy for (a) single modality tuning, (b) multi-task tuning, and (c) our newly proposed VELoRA for RGB-Event effective fusion.}
\label{fig:firstIMG}
\end{figure*}

Inspired by the success of LoRA for the large language model, as shown in Fig.~\ref{fig:firstIMG}, in this paper, we propose a novel Parameter Efficient Fine-Tuning strategy for RGB-Event based pattern recognition, termed VELoRA. Given the RGB frames and event streams, we first obtain the frame difference of the inputs and capture the motion information using a frame difference backbone network. Then, we propose to encode the RGB and event data with the newly proposed modality-specific LoRA tuning based on pre-trained vision foundation models. To boost the interactions of the dual modalities, we reconstruct the modality features using the other one. After that, we concatenate the RGB, event, and frame difference features and feed them into a fusion network with modality-shared LoRA tuning. The output will be concatenated and mapped into detailed categories with a classification head. Extensive experiments on multiple event-based pattern recognition fully validated the effectiveness of our proposed framework.

To sum up, we draw the contributions of this paper as the following three aspects: 

$\bullet$ We propose a novel parameter-efficient fine-tuning strategy for RGB-Event-based pattern recognition, termed VELoRA, which achieves a better tradeoff between the computational cost and the accuracy. 

$\bullet$ We exploit the interactions between the dual modalities using frame difference and reconstruction schemes for the RGB-Event-based pattern recognition. 

$\bullet$ Extensive experiments on multiple event-based pattern recognition benchmark datasets fully validated the effectiveness of our proposed PEFT strategy.

\textit{The rest of this paper is organized as follows:} 
We introduce the related works in Section~\ref{sec::relatedWorks} with a focus on event-based recognition, RGB-Event-based recognition, and Parameter Efficient Fine-Tuning. In Section~\ref{sec::methods}, we will describe the details of our newly proposed VELoRA framework and validate its effectiveness in Section~\ref{sec::experiments}. Finally, we conclude this paper and propose possible future works in Section~\ref{sec::conclusion}.

\section{Related Works} \label{sec::relatedWorks}

In this section, we review the related works on Event-based recognition, RGB-Event-based recognition, and Parameter-Efficient Fine-Tuning strategies. More related works can be found in the following surveys~\cite{gallego2020eventSurvey} and paper list~\footnote{\url{https://github.com/Event-AHU/Event_Camera_in_Top_Conference}}.

\noindent $\bullet$ \textbf{Event based Recognition.~}  
Event-based recognition can be typically categorized into 
CNN-based~\cite{WangDSWZS019}, 
GNN-based~\cite{Xie2022VMVGCNVM}~\cite{schaefer2022aegnn}, 
Transformer~\cite{yang2024uncertainty}~\cite{yuan2023learning}, 
and SNN-based~\cite{zhou2022spikformer}~\cite{fang2021incorporating} approaches. 
Chen et al.~\cite{chen2020dynamic} proposes a novel concept for 
gesture recognition by taking events as three-dimensional points
data and employed into a dynamic graph CNN.
Zhu et al.~\cite{zhu2018ev} proposes an novel approach that employs
Time Surfaces to distill spatiotemporal characteristics from event-based data.
Wang et al. \cite{wang2019ev} presents a gait recognition method named EV-Gait, which is based on Convolutional Neural Networks(CNNs). it addresses the challenge of noise in event streams by enforcing motion consistency.
Graph Neural Networks (GNNs) are increasingly being utilized to analyze event data, with notable models such as the Asynchronous Event-based Graph Neural Network (AEGNN) ~\cite{schaefer2022aegnn}, which treats event data as dynamic spatio-temporal graphs. it preserves sparsity and high temporal resolution by selectively updating nodes that are directly affected by incoming events while reducing computational complexity and latency. 
Xie et al.~\cite{Xie2022VMVGCNVM} introduced VMV-GCN, a voxel-centric geometric learning framework aimed at consolidating multi-view volumetric information.
In addition, Spiking Neural Networks (SNNs), which mimic the behavior of biological neurons, provide a more energy-efficient solution. This is exemplified by models such as Spikformer~\cite{zhou2022spikformer}, which integrates spiking neurons with Transformer architectures. 
Xiao et al. introduced the HMAX Spiking Neural Network (HMAX SNN)~\cite{xiao2019event}, utilizing multi-spike encoding to capture temporal features. 
Liu et al. proposed Motion SNN~\cite{liu2021event}, which incorporates motion information into a multilayer SNN framework.
Lee et al.~\cite{lee2016training} developed an event camera recognition method based on Spiking Neural Network(SNNs),
leveraging computationally inspired supervised learning techniques, including backpropagation.

In addition, The SpikMamba~\cite{chen2024spikmamba} framework, as introduced by Chen et al., is designed to harness the energy efficiency of spiking neural networks(SNNs) and the long sequence modeling capability of Mamba.
this integration is particularly effective for capturing global features from event data that is spatially sparse.
Wang et al.~\cite{wang2024event} propose event-based Human Action Recognition(HAR) model Known as EVMamba, representing a significant advancement in the field by integrating a spatial plane multi-directional scanning mechanism along with an innovative voxel temporal information.
Yang et al.~\cite{yang2024uncertainty} have developed a novel Mobile-Former network that emphasizes uncertainty-aware information
propagation for the purpose of pattern recognition, which combines
the MobileNet and Transformer network.


\noindent $\bullet$ \textbf{RGB-Event based Recognition} 
Multi-modal-based research has been widely exploited in many downstream tasks~\cite{wang2023unleashing, wang2022mfgnet}. RGB-Event-based recognition leverages complementary information from conventional RGB frames and event data to enhance pattern recognition, particularly in dynamic environments. Several studies have explored architectures that effectively integrate these modalities to harness their individual strengths. 
For instance, Huang et al.~\cite{9897492} proposed a cross-attention module that facilitates cross-modal feature selection, enabling the model to dynamically prioritize relevant information from both RGB and event inputs. 
Wang et al. introduced SSTFormer~\cite{wang2023sstformer}, a memory-supported Transformer framework that encodes RGB frames while using a spiking neural network (SNN) to process raw event streams. This approach balances detailed spatial information with the temporal dynamics inherent in event data. 
Additionally, Li et al.~\cite{li2023semantic} developed the SAFE model, which incorporates semantic labels alongside RGB and event data through a semantic-aware frame-event fusion mechanism, enhancing recognition accuracy by leveraging large-scale vision-language pre-training. 
TSCFormer~\cite{wang2023unleashing} is proposed by aggregating the temporal shift-based convolutional neural networks (CNNs) and lightweight Transformer modules for efficient modality fusion. 
Different from these works, in this paper, we achieve effective RGB-Event fusion via the modality-shared LoRA tuning scheme.

\noindent $\bullet$ \textbf{Parameter-Efficient Fine-Tuning.~}  
Parameter-Efficient Fine-Tuning (PEFT), initially introduced in the field of natural language, adjusts only a subset of the parameters of a pre-trained model instead of the entire set, to reduce computational resource consumption and training time while maintaining or improving model performance. The Low-Rank Adaptation (LoRA) is proposed to incorporate trainable low-rank matrices into the transformer architecture, which serves to approximate weight adjustments~\cite{hu2021lora}. Although LoRA blocks are efficient,
the rank is fixed and cannot be altered. To address this, DyLoRA~\cite{valipour2022dylora} introduces a dynamic approach to
low-rank adaptation, instead of training LoRA block for a single.
Zhang et al.~\cite{zhang2023lora} have introduced LoRA-FA, freezing the projection-down weight of matrix A, while allowing the projection-up weight of matrix B to be updated, it reduces training parameters and resource consumption without performance degradation.
However, the fine-tuning performance is not ideal due to the varying importance of different weight parameters. AdaLoRA~\cite{zhang2023adalora} introduces a novel allocation of parameter budgets based on a calculated importance score.
ALoRA~\cite{liu2024alora} considers that a fixed intrinsic rank may not be flexible enough for various downstream tasks, allowing dynamic adjustment of the rank during the training process, which can gradually prune away ranks that have a negative impact. 
AdaptFormer~\cite{chen2022adaptformer} introduces lightweight adapters, which can enhance the transferability of the Vision Transformer (ViT) without updating its original pre-trained parameters. It marks the first instance of adapting vision Transformers to a wide range of downstream visual recognition tasks using adapters.

In this work, we adopt a parameter-efficient fine-tuning strategy that leverages Low-Rank Adaptation (LoRA) blocks to enhance Transformer layers in both modality-specific and modality-shared ways. This architecture utilizes modality-specific LoRA tuning in the lower-level Transformer blocks, which are tailored to handle the unique characteristics of RGB and event inputs. This enables the model to efficiently capture spatiotemporal features from each modality, without updating the full set of parameters, thereby reducing computational overhead.


\begin{figure*}
\centering
\includegraphics[width=\textwidth]{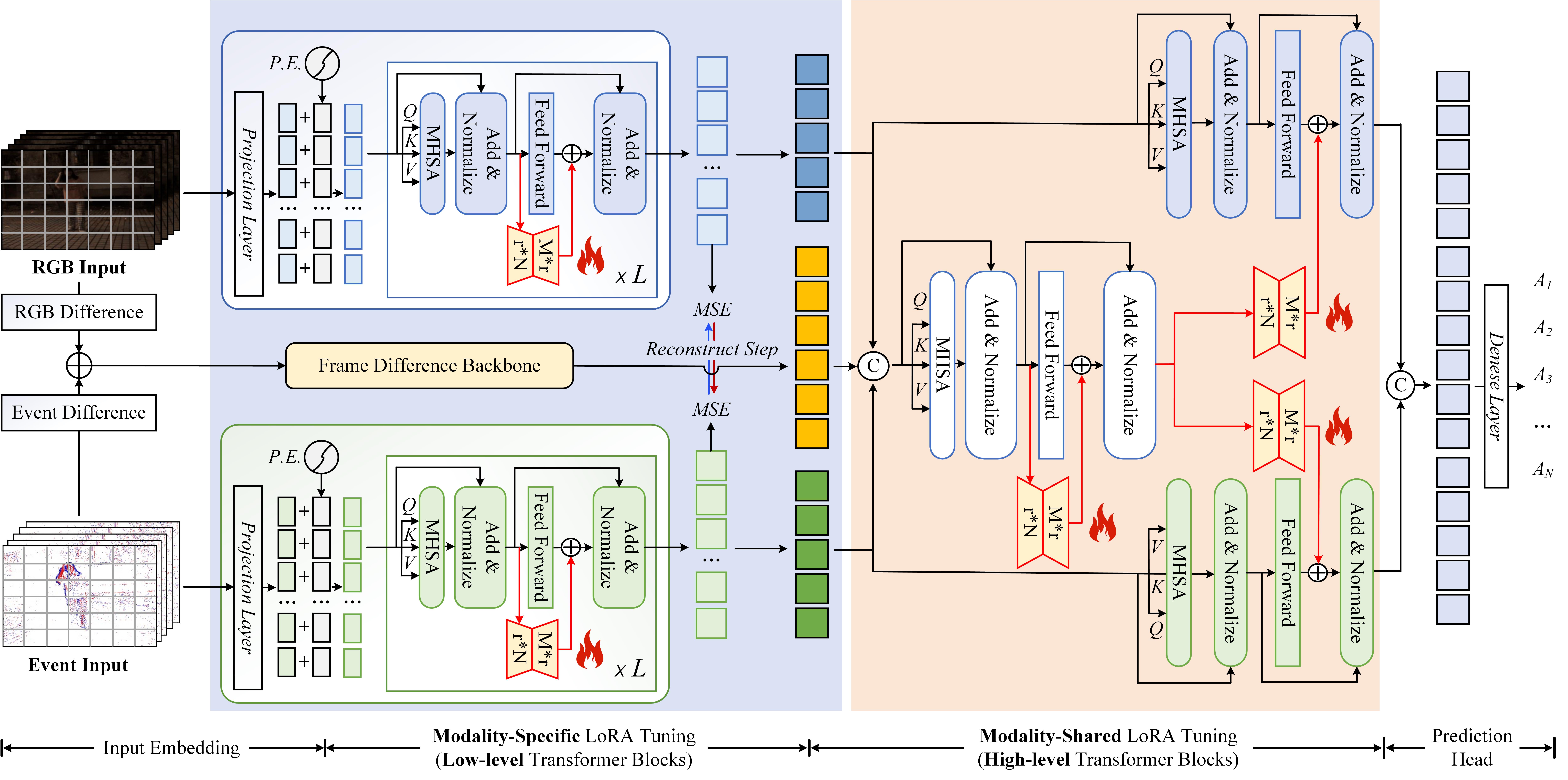}
\caption{\textbf{An overview of our proposed Low-Rank Adaptation Approach for Efficient Visible-Event Pattern Recognition, termed VELoRA.}
We introduce a novel fine-tuning approach that integrates modality-specific and shared components, enabling the model to preserve sensitivity to distinct modalities while also extracting shared information across them, which boosts performance on multimodal tasks. We designate the last block as the high-level Transformer block, with the remaining blocks functioning as low-level ones. For RGB and event inputs, we encode them using a pre-trained large vision model and introduce a reconstruction loss to enhance feature fusion and generalization. Additionally, we incorporate the differences between consecutive frames as an auxiliary modality in the modality-specific stage to aid RGB and event modalities. In the shared modality stage, we refine the original output with the combined tri-modal data. The refined features are then fed into the classification head for final categorization. 
}
\label{fig:framework}
\end{figure*}

\section{Our Proposed Approach}  \label{sec::methods}

In this section, we will first give a preliminary introduction to the LoRA strategy, and then introduce the overview of our proposed VELoRA-based recognition framework. After that, we focus on the network architecture and loss function.

\subsection{Preliminary: LoRA Strategy}
Low-Rank Adaptation (LoRA) is an efficient fine-tuning technique designed to enhance model performance by incorporating trainable low-rank matrices into pre-trained models. The central premise of LoRA is to maintain the integrity of the pre-trained model's weights while inserting trainable layers within each Transformer block. This approach dramatically reduces the number of parameters that need to be updated for downstream tasks, consequently lowering the GPU memory requirements. During the LoRA fine-tuning phase, the method prioritizes the linear layer within the Transformer block, achieving fine-tuning quality that is on par with full parameter fine-tuning, yet it is computationally more efficient and requires fewer resources. This attribute confers a significant practical advantage to the LoRA fine-tuning method.

LoRA emulates the effects of full-parameter fine-tuning by decomposing the weight matrix into two smaller matrices, which are then used to approximate the original matrix. The original matrix is kept frozen, and it is decomposed into matrices $\textbf{A}$ and $\textbf{B}$, where the rank $r$ is much smaller than the dimension $d$. Matrix $\textbf{A}$ is initialized with values drawn from a random Gaussian distribution, while matrix $\textbf{B}$ is initialized as a matrix of zeros. Given the input feature $x$, the compute procedure can be written as: 
\begin{equation}
h = \textbf{W}_0x + \Delta \textbf{W}x, \quad \Delta \textbf{W} = \textbf{B}\textbf{A}
\end{equation}
Here, $\textbf{W}_0$ represents the pre-trained weights that remain frozen throughout the training process. The matrices $\textbf{B}$ and $\textbf{A}$ are the ones that will be continuously optimized during training.

Although the LoRA model has these advantages, there has been no work attempting to apply this strategy to the RGB-Event multi-modal classification task. This paper attempts to propose a multi-modal fine-tuning strategy based on LoRA in the hope of achieving a better balance between the cost of model fine-tuning and recognition performance.

\subsection{Overview} 
As depicted in Fig.~\ref{fig:framework}, our proposed VELoRA framework takes RGB frames and event streams as inputs and employs pre-trained Large Vision Models (LVM) to encode the RGB frames and event streams independently, without parameter sharing. We consider the Transformer layers close to the input end as extracting low-level features, while the layers near the output end are seen as high-level features. We conduct modality-specific LoRA tuning on the low-level Transformer layer to obtain feature representations specific to each modality. Meanwhile, we also extract the frame difference information from the visible light and event stream to obtain motion cues. The outputs are concatenated and fed into a modality-shared LoRA tuning module to achieve multi-modal feature fusion and interaction. Finally, we adopt a classification head to predict the category labels. The VELoRA model designed for this multi-modal fusion framework achieves high precision while maintaining a low number of parameters that require adjustment.

\subsection{Network Architecture}


$\bullet$ \textbf{Input Encoding.} 
Given the input image set $\mathcal{T}_v = \{v_1, v_2, \ldots, v_M\}$ and the event stream $\mathcal{T}_e = \{e_1, e_2, \ldots, e_N\}$, where $M$ and $N$ represent the number of video frames and event points, respectively. Each event point in the stream is defined as $e_i = [x, y, t, p]$, with $(x, y)$ indicating spatial coordinates, $t$ the timestamp, and $p$ the polarity. Initially, we convert the event stream $\mathcal{E}$ into event images aligned with the timestamps of the video frames. To effectively capture the feature representations of these inputs, we employ the large vision model $LVM$ (ViT-B/16-based CLIP model~\cite{rao2022denseclip}) for encoding, striking a balance between accuracy and efficiency. The encoded features $\mathcal{F}_v \in \mathbb{R}^{197 \times 768}$ and $\mathcal{F}_e \in \mathbb{R}^{197 \times 768}$ are expressed as:
\begin{equation}
\mathcal{F}_v = LVM(\mathcal{T}_v), \quad \mathcal{F}_e = LVM(\mathcal{T}_e). 
\end{equation}

To better capture motion information while reducing background interference, we introduce frame difference to further improve our model, i.e., $\mathcal{D} = \{d_1, d_2, \ldots, d_{C-1}\}$, where $C$ represents the total number of frames (we set it as 8 for this study). We calculate the mean of the differences between consecutive frames and process it through the aforementioned vision backbone network, 
\begin{equation}
\mathcal{F}_d = LVM(\mathcal{T}_d). 
\end{equation}
Here, $\mathcal{F}_d \in \mathbb{R}^{197 \times 768}$, 197 and 768 indicate the number of tokens and the dimension of each token, respectively.

If the large vision backbone is trained using full fine-tuning, it can achieve good accuracy, but it is followed by high computational complexity. To achieve a better balance between the two aspects, this paper attempts to design a novel RGB-Event lightweight fine-tuning method called VELoRA. Our VELoRA primarily consists of two major modules: one for modality-specific and another for modality-shared LoRA tuning. We utilize the last block as our high-level Transformer block, while the remaining blocks serve as low-level blocks for fine-tuning specific modalities. We will introduce this module in detail in the remaining paragraphs.

$\bullet$ \textbf{Modality-specific LoRA Tuning.~}  
To achieve efficient fine-tuning, we propose the modality-specific LoRA tuning strategy and employ the same backbone network for processing RGB, event, and frame difference features. Note that all parameters within each Transformer block are frozen. In each branch, we exclusively update the weights of the LoRA in the MLP linear layers of the Transformer blocks. We also introduce a reconstruction step between the RGB and Event branches to further boost the interactions of the dual modalities. A Transformer layer is adopted to reconstruct the modality features using another one. We apply MSE (Mean Squared Error) loss to regularize the transformation between these modalities. 
Thus, for the RGB frame branch, the computing procedure in a Transformer block using Modality-specific LoRA Tuning can be summarized as:
\begin{equation}
\mathcal{F}_{v}^{(l+1)} = \mathcal{F}_{v}^{(l)} + (\textbf{W}_0 + \Delta \textbf{W}) \mathcal{F}_{v}^{(l)}
\end{equation}
Here, $l$ represents the $l$-th layer of the Transformer block, $\textbf{W}$ denotes the pre-trained weights, and $\Delta \textbf{W}$ signifies the product of the weights in the $\textbf{B}$ and $\textbf{A}$ matrices. The other two branches follow a similar calculation method, which will not be elaborated here.


$\bullet$ \textbf{Modality-shared LoRA Tuning.~}  
After obtaining modality-specific feature representations, the key to multi-modal tasks lies in how to further facilitate the interaction and fusion of features across different modalities. This paper proposes a novel multi-modal fine-tuning strategy called Modality-shared LoRA Tuning, which accomplishes modal interaction and fusion through lightweight fine-tuning. Specifically, we first concatenate the multi-modal features and feed them into a Transformer layer for fusion. We refine the multi-modal features $\mathcal{F}_{fuse}$ using LoRA, i.e., 
\begin{equation}
\mathcal{F}_{fuse} = LoRA(MHSA([F_v, F_e, F_d]))
\end{equation} 
where $MHSA$ is short for Multi-Head Self-Attention, [,] denotes the concatenate operation. 
Then feed these features into two LoRA branches, where under the guidance of the multi-modal features, we further learn the modality-specific feature representations, i.e., 
\begin{equation}
\bar{\mathcal{F}_{v}} = \mathcal{F}_{v} + LoRA(\mathcal{F}_{fuse}) 
\end{equation} 
Then, the residual and normalization operations are applied to the enhanced feature for the Event image branch. We adopt similar operations for the RGB image branch. Finally, we concatenate the enhanced RGB and Event features and send them to the classification head for the final classification.

\subsection{Loss Function} 
In our study, we employ the cross-entropy loss function, a standard choice for classification tasks, to quantify the discrepancy between the model's predicted probability distribution and the actual distribution of the true labels. The loss function is defined as:
\begin{equation}
\mathcal{L}(y, \hat{y}) = -\sum_{i=1}^Ny_i\log(\hat{y}_i)
\end{equation}
where $N$ refers to the number of classes, and $y$ and $\hat{y}$ denote the ground truth and predicted labels, respectively.

To generate higher-quality images by reducing noise and blurring, and to facilitate cross-modal information exchange, we focus on minimizing the reconstruction loss between RGB images and event streams~\cite{wang2024top}. This is achieved through the use of the Mean Squared Error (MSE) loss function in our reconstruction phase. The formulas for the reconstruction losses are as follows:
\begin{equation}\mathcal{L}_{RTE} = \frac{1}{M}\sum_{i=1}^M(y_e-\hat{y_e})^2\end{equation}
\begin{equation}\mathcal{L}_{ETR} = \frac{1}{M}\sum_{i=1}^M(y_r-\hat{y_r})^2\end{equation}
Here, $M$ represents the number of the patch tokens, $y_e$ and $\hat{y_e}$ are the raw and reconstructed features for the event stream, while $y_r$ and $\hat{y_r}$ are the original and reconstructed features for the RGB image.  
Therefore, the composite loss of our model is the sum of these individual losses, expressed as:
\begin{equation}\mathcal{L}_{total} = \mathcal{L}_{RTE}+\mathcal{L}_{ETR}+\mathcal{L}(y, \hat{y}) \end{equation}

\section{Experiments} \label{sec::experiments}

\subsection{Datasets and Evaluation Metric} 
In this part, we primarily conduct our experiments on two event-based classification datasets, namely HARDVS~\cite{wang2024hardvs} and PokerEvent~\cite{wang2023sstformer}. A brief introduction to the two datasets is given below. 

\noindent $\bullet$ \textbf{PokerEvent dataset.~} 
The PokerEvent dataset is mainly composed of character patterns in poker cards, which consists of 114 classes and includes 24,415 RGB-Event samples, then we divide it into 16216 and 8199 for training and testing respectively.

\noindent $\bullet$ \textbf{HARDVS dataset.~}
This dataset primarily consists of human activities, such as walking, running, and other similar actions, which are collected through a DVS346 event camera, and it is composed of 300 classes, which we have divided into 64522 for training and 32386 for testing.

To test our method and compare it with other state-of-the-art methods, we adopt the \textit{top-1} accuracy as the evaluation metric.

\subsection{Implementation Details} 
We train the model for a total of 50 epochs on the PokerEvent databases and the HARDVS database, using a batch size of 4. The initial learning rate was set to 10e-04, and the CosineLRScheduler was utilized to dynamically adjust the learning rate during the training phase. 
For optimizing our network, we employ the AdamW optimizer~\cite{loshchilov2017decoupled}. Our model architecture is based on ViT-Base, with all parameters remaining frozen throughout the training, except for the weights of LoRA, which were optimized. The rank \textit{r} was set to 4, and only the MLP (Multi-Layer Perceptron) layers within each Transformer block were updated. The experiment is conducted on a server equipped with NVIDIA GeForce RTX 3090 GPUs. More details can be found in our source code.

\subsection{Comparison on Public Benchmark Datasets}  

\begin{table}[]
\centering
\caption{Results on the PokerEvent dataset.} 
\label{results_PokerEvent}
\begin{tabular}{l|l|l|c}
\hline
\textbf{Algorithm}    & \textbf{Publication}     & \textbf{Backbone}       & \textbf{Precision} \\ 
\hline
C3D~\cite{tran2015learning}          & ICCV-2015  & 3D-CNN         & 51.76     \\
TSM~\cite{lin2019tsm}          & ICCV-2019  & ResNet-50      & 55.43     \\
TAM~\cite{liu2021tam}          & ICCV-2021  & ResNet-50      & 53.65     \\
ACTION-Net~\cite{wang2021action}   & CVPR-2021  & ResNet-50      & 54.29     \\
V-SwimTrans~\cite{liu2022video}  & CVPR-2022  & Swin-T.Former & 54.17     \\
TimeSformer~\cite{bertasius2021space}  & ICML-2021  & ViT-B/16       & 55.69     \\
X3D~\cite{feichtenhofer2020x3d}          & CVPR-2020  & ResNet         & 51.75     \\
MVIT~\cite{li2022mvitv2}         & CVPR-2022  & ViT            & 55.02     \\
SSTformer~\cite{wang2023sstformer}    & arXiv-2023 & SNN-Former     & 54.74     \\
SAFE~\cite{li2023semantic}        & PR-2024          & ViT-B/16       & 57.63     \\ 
\hline
VELoRA (Ours) &-            & ViT-B/16       & \textbf{57.99}     \\ 
\hline
\end{tabular}
\end{table}

\noindent $\bullet$ \textbf{Results on PokerEvent Dataset.~} 
As depicted in Table~\ref{results_PokerEvent}, we benchmark our proposed model against several robust classical models, including C3D~\cite{tran2015learning}, TSM~\cite{lin2019tsm}, TAM~\cite{liu2021tam}, as well as a selection of Transformer-based models such as TimeSformer~\cite{bertasius2021space}, V-SwimTrans~\cite{liu2022video}, and MVIT~\cite{li2022mvitv2}.
The comparative results indicate that our model achieves overall superior performance compared to the baseline models. In detail, our model attains an accuracy of 57.99\%, marginally surpassing the state-of-the-art SAFE~\cite{li2023semantic}, which shares the same backbone architecture as our model and outperforms all other models by a clear margin. The impressive results on this dataset comprehensively validate the efficacy of our model in PokerEvent recognition and substantiate the utility of our proposed method.

\noindent $\bullet$ \textbf{Results on HARDVS Dataset.~} 
The HARDVS benchmark dataset offers both RGB frames and event streams for the purpose of human activity recognition. In this section, we present the experimental results for both the RGB image and event stream modalities. As indicated in Table~\ref{results_HARDVS}, our model has achieved an accuracy rate of 50.89\%, outperforming several prominent models, including slowFast~\cite{feichtenhofer2019slowfast} (46.54\%), X3D~\cite{feichtenhofer2020x3d} (47.38\%), ACTION-Net~\cite{wang2021action} (46.85\%), ResNet18~\cite{he2016deep} (49.20\%), and R2PlusID~\cite{tran2018closer} (49.06\%), and is slightly higher than ESTF~\cite{wang2024hardvs} (49.93\%), SAFE~\cite{li2023semantic} (50.17\%), and C3D~\cite{tran2015learning} (50.88\%). These results underscore that our VELoRA model attains comparable, if not superior, performance to the state-of-the-art models, thereby fully validating the efficacy of our proposed RGB-Event recognition framework. Our work on the HARDVS dataset offers a valuable contribution to the research community and provides practical insights into the promising potential of our model in the field of human activity recognition.

\begin{table}[]
\centering
\caption{Results on the HARDVS dataset.} 
\label{results_HARDVS}
\begin{tabular}{l|l|l|c} 
\hline
\textbf{Algorithm}    & \textbf{Publication}     & \textbf{Backbone}       & \textbf{Precision} \\  
\hline
C3D~\cite{tran2015learning}                     & ICCV-2015    & 3D-CNN     & 50.88     \\
ResNet18~\cite{he2016deep}                      & CVPR-2016    & ResNet18   & 49.20     \\
ACTION-Net~\cite{wang2021action}                & CVPR-2021    & ResNet-50  & 46.85     \\
SlowFast~\cite{feichtenhofer2019slowfast}       & ICCV-2019    & ResNet-50  & 46.54     \\
R2Plus1D~\cite{tran2018closer}                  & CVPR-2018    & ResNet-34  & 49.06     \\
X3D~\cite{feichtenhofer2020x3d}                 & CVPR-2020    & ResNet     & 47.38     \\
ESTF~\cite{wang2024hardvs}                      & AAAI-24      & ResNet-18  & 49.93     \\
SAFE~\cite{li2023semantic}                      & PR-2024      & ViT-B/16   & 50.17     \\ 
\hline
VELoRA (Ours) &-                 & ViT-B/16  & \textbf{50.89}     \\ 
\hline
\end{tabular}
\end{table}

\subsection{Ablation Study} 

\noindent $\bullet$ \textbf{Component Analysis.~} 
In this part, we assess the impact of various components within our model, including the frame difference module, the reconstruction step, Modality-specific LoRA Tuning, and Modality-shared LoRA Tuning, primarily by evaluating their recognition performance on the PokerEvent datasets. As depicted in Table~\ref{results_CA}, when the model was trained with only the frame difference module or the reconstruction step, the highest accuracy achieved was 55.94\%. However, when both modules were employed concurrently, the model's accuracy improved to 56.79\%. We posit that the RGB and event stream reconstruction loss aids in facilitating the exchange of information between different modalities, thereby enhancing the model's ability to understand and leverage information from each modality.

Initially, we employed full fine-tuning for model training, acknowledging the considerable memory and resource consumption associated with this approach. Subsequently, we investigated the effects of specific LoRA fine-tuning and shared LoRA fine-tuning on the experimental results. Notably, even with a reduction in the number of parameters, our model's accuracy surpasses that of full fine-tuning, reaching 56.97\% and 57.12\% respectively. The specific LoRA fine-tuning is designed for individual modalities, while the shared LoRA fine-tuning further augments the features across multiple modalities. In an attempt to harness the benefits of both methods, we combined the fine-tuning approaches. Remarkably, this led to the model achieving the highest accuracy of 57.99\%. The experimental results suggest that each module contributes to the overall recognition performance, albeit to varying degrees.

\begin{table}
\caption{Component Analysis of Our Framework on PokerEvent Dataset.}   
\label{results_CA} 
\resizebox{\columnwidth}{!}{
\begin{tabular}{c|cccc|c} 		
\hline 
\textbf{No.}  & \textbf{F.D.} &\textbf{Reconst.}&\textbf{LoRA-Spe}&\textbf{LoRA-Share}  &\textbf{Results}   \\   
\hline 
1 &\cmark       &           &           &           &55.78    \\
2 &             &\cmark     &           &           &55.94    \\
3 &\cmark       &\cmark     &           &           &56.79    \\
4 &\cmark       &\cmark     &\cmark     &           &56.97    \\
5 &\cmark       &\cmark     &           &\cmark     &57.12    \\
6 &\cmark       &\cmark     &\cmark     &\cmark     &57.99    \\
\hline 
\end{tabular}} 
\end{table}

\begin{figure*}
\centering
\includegraphics[width=\textwidth]{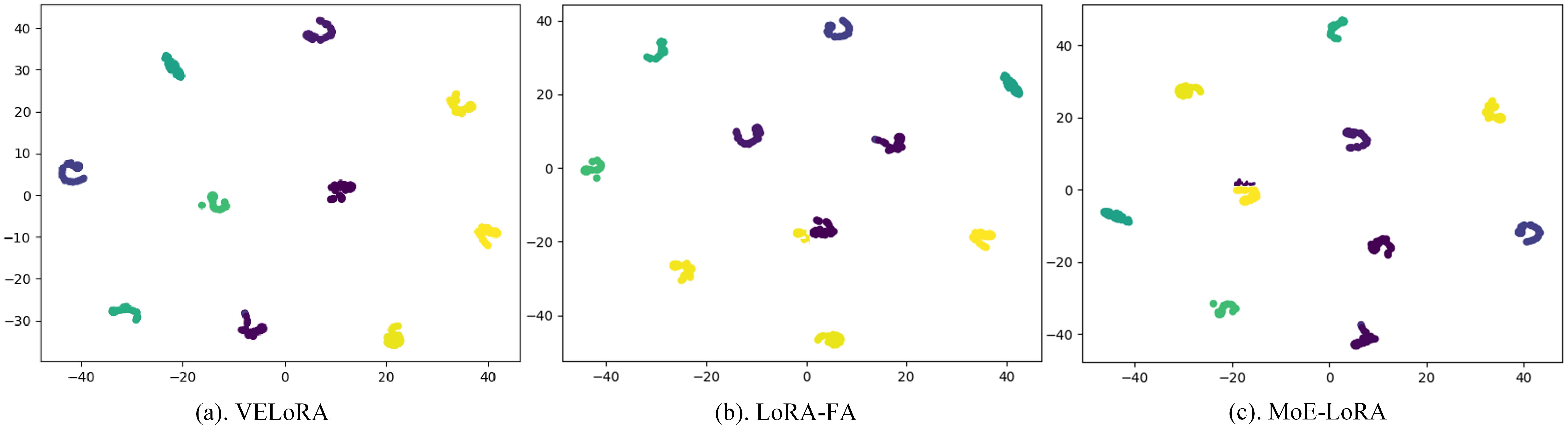}
\caption{Visualization of feature distribution of (a) Ours, (b) full fine-tuning on PokerEvent.} 
\label{fig:enter-label}
\end{figure*}

\noindent $\bullet$ \textbf{Analysis of the Number of Input Frames.~} 
In this sub-section, we examine the influence of the quantity of input frames on the precision of our model. We trained the model using 4, 5, 6, 8, and 10 frames as inputs, respectively, and compared the training outcomes, as presented in Table~\ref{resultsDiffFrames}. Contrary to expectations, our findings did not show a steady improvement in the model's accuracy with an increase in the number of input frames. The model trained with 8 input frames yielded the highest accuracy rate of 57.99\%, while the model trained with only 4 frames achieved the lowest rate of 55.02\%. Surprisingly, the model trained with 10 input frames did not outperform the one trained with 8 frames, reaching an accuracy of 57.54\%. A potential reason for this trend could be the existence of an ideal limit that further increases the number of frames fails to offer extra valuable visual insights and might bring in noise. Another reason might be that a high number of frames could add a layer of dynamic complexity that could negatively impact the model's efficiency. 

In summary, the quantity of input frames is an essential consideration when training neural networks for video analysis. Increasing the number of input frames moderately can boost precision, but an excessive amount could lead to noise and demand more computational resources as well as longer training periods.

\begin{table}[!htp]     
\caption{Experimental results of different input frames.}   
\label{resultsDiffFrames} 
\resizebox{\columnwidth}{!}{
\begin{tabular}{c|c|c|c|c|c} 		
\hline 
\textbf{Frames}    & \textbf{4}    &\textbf{5}   &\textbf{6}    &\textbf{8}  &\textbf{10}   \\   
\hline 
Results            &55.02          & 57.22       & 56.75          &57.99  &57.54   \\
\hline 
\end{tabular}} 
\end{table}

\noindent $\bullet$ \textbf{Analysis on Places and Times of LoRA.~} 
In this part, we investigate the influence of the LoRA rank $r$ and its placement within the transform blocks, as detailed in Table~\ref{rank r}. Unexpectedly, our findings did not reveal a monotonic improvement in the model's accuracy with an increase in the rank $r$. In fact, the model trained with a rank of 4 achieved the highest accuracy rate of 57.99\%, while the model with a rank of 8 had the lowest rate at 55.95\%. Similarly, the model trained with a rank of 12 did not perform as well as the one with a rank of 4, registering an accuracy of 57.08\%. However, this was slightly better than the accuracy of 56.35\% achieved with a rank of 6.

Furthermore, we explored the effect of the placement of LoRA within the network. Our focus was primarily on the QKV (Query, Key, Value) and the linear layers of the MLP (Multi-Layer Perceptron), as shown in Table \ref{results_loraPOS}. We observed that applying LoRA exclusively to the QKV resulted in an accuracy of 56.69\%. Augmenting this with the MLP linear layers improved the model's accuracy to 57.26\%. This led us to hypothesize that incorporating LoRA in the MLP layers is beneficial for the model. To test this, we conducted a separate experiment applying LoRA only to the MLP layers, which surprisingly yielded the best accuracy of 57.99\%. The increase in rank, coupled with the expansion of LoRA application positions, necessarily leads to an increase in the number of training parameters and memory consumption. Consequently, both the rank $r$ and the positions where LoRA is applied are critical parameters in the training process.

\begin{table}[!htp]  
\small     
\centering 
\caption{Results with different rank $r$}  
\label{rank r} 
\begin{tabular}{c|c|c|c|c} 		
\hline 
\textbf{r}  & \textbf{4} &\textbf{6} &\textbf{8}    &\textbf{12}   \\   
\hline 
Results &57.99   & 56.35    &55.95            &57.08     \\
\hline 
\end{tabular}
\end{table}

\begin{table}[!htp]
\small     
\centering  
\caption{Results of different layers for LoRA inserting.}  
\label{results_loraPOS} 
\begin{tabular}{c|c|c|c} 		
\hline 
\textbf{Location}  & \textbf{QKV} &\textbf{MLP} &\textbf{QKV+MLP}      \\   
\hline 
Results     &56.69   & 57.99   &57.26                \\
\hline 
\end{tabular}
\end{table}

\begin{figure*}
\centering
\includegraphics[width=\textwidth]{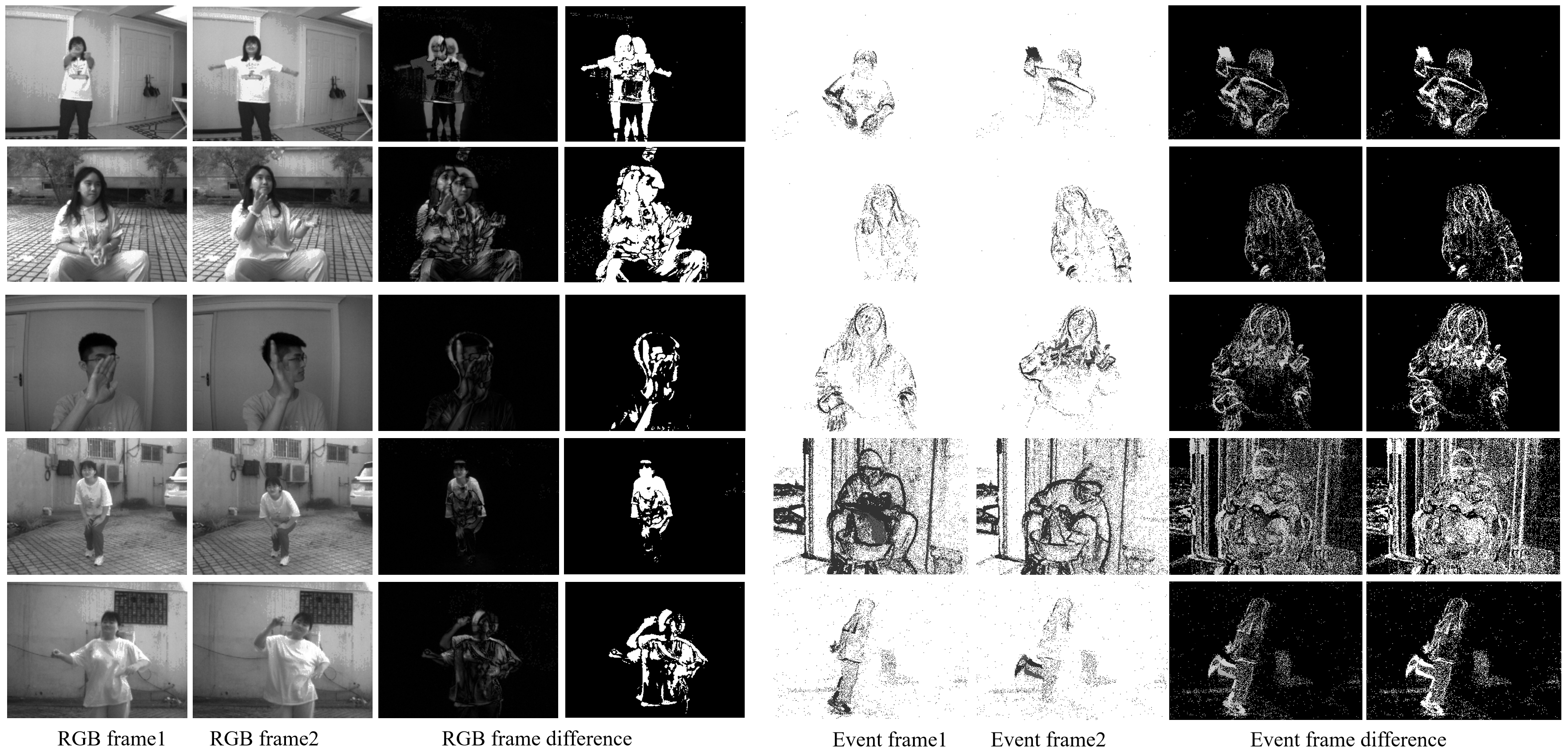}
\caption{Visualization of the  RGB frame differences (left) and Event frame differences (right) on the HARDVS dataset.} 
\label{fig:frameDiff_VIS}
\end{figure*}

\begin{figure*}
\centering
\includegraphics[width=1\linewidth]{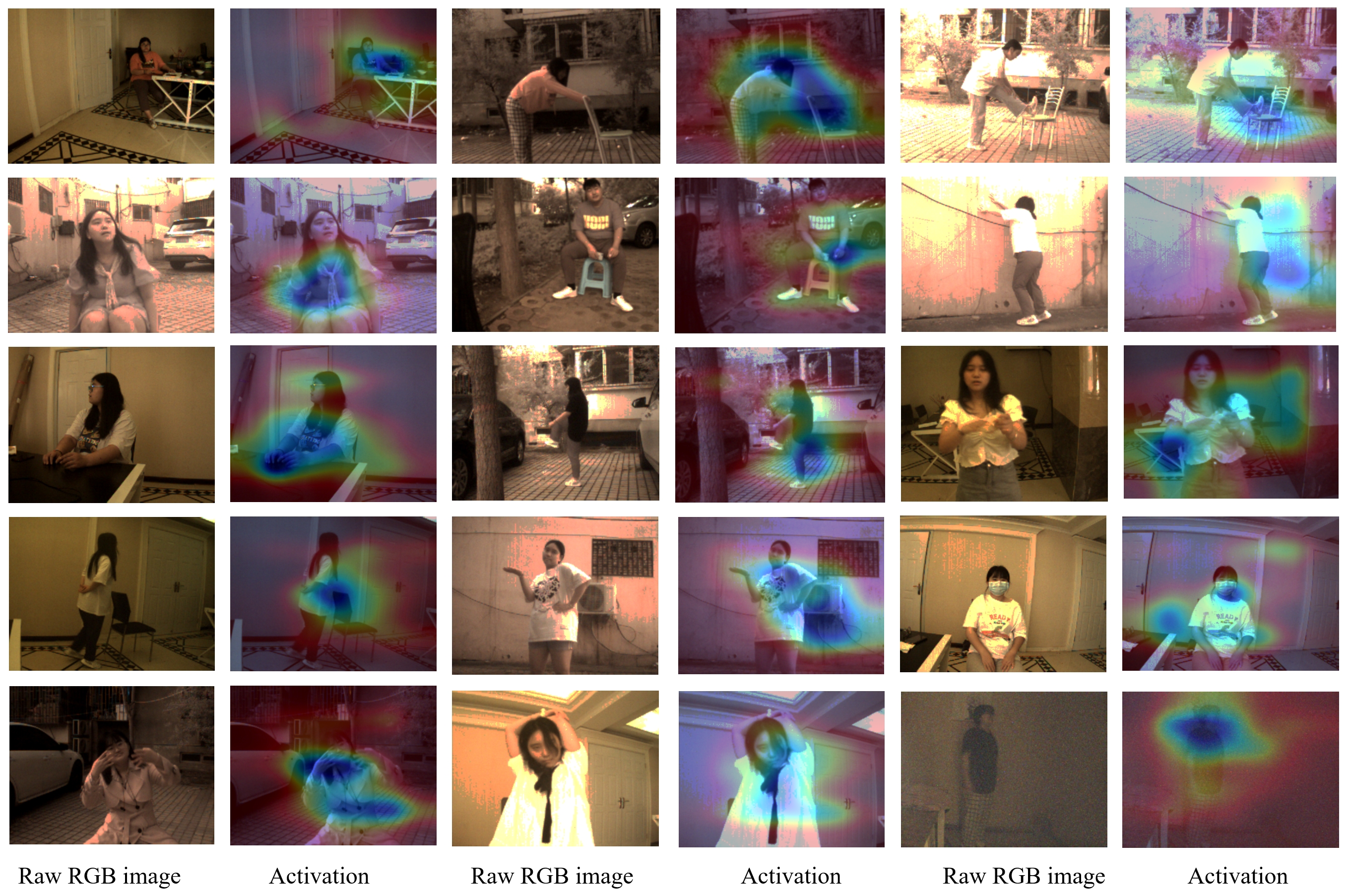}
\caption{Visualization of the raw RGB image and activation maps. Note that, the blue denotes a higher activation response.} 
\label{fig:activation_VIS}
\end{figure*}

\begin{figure*}
\centering
\includegraphics[width=\textwidth]{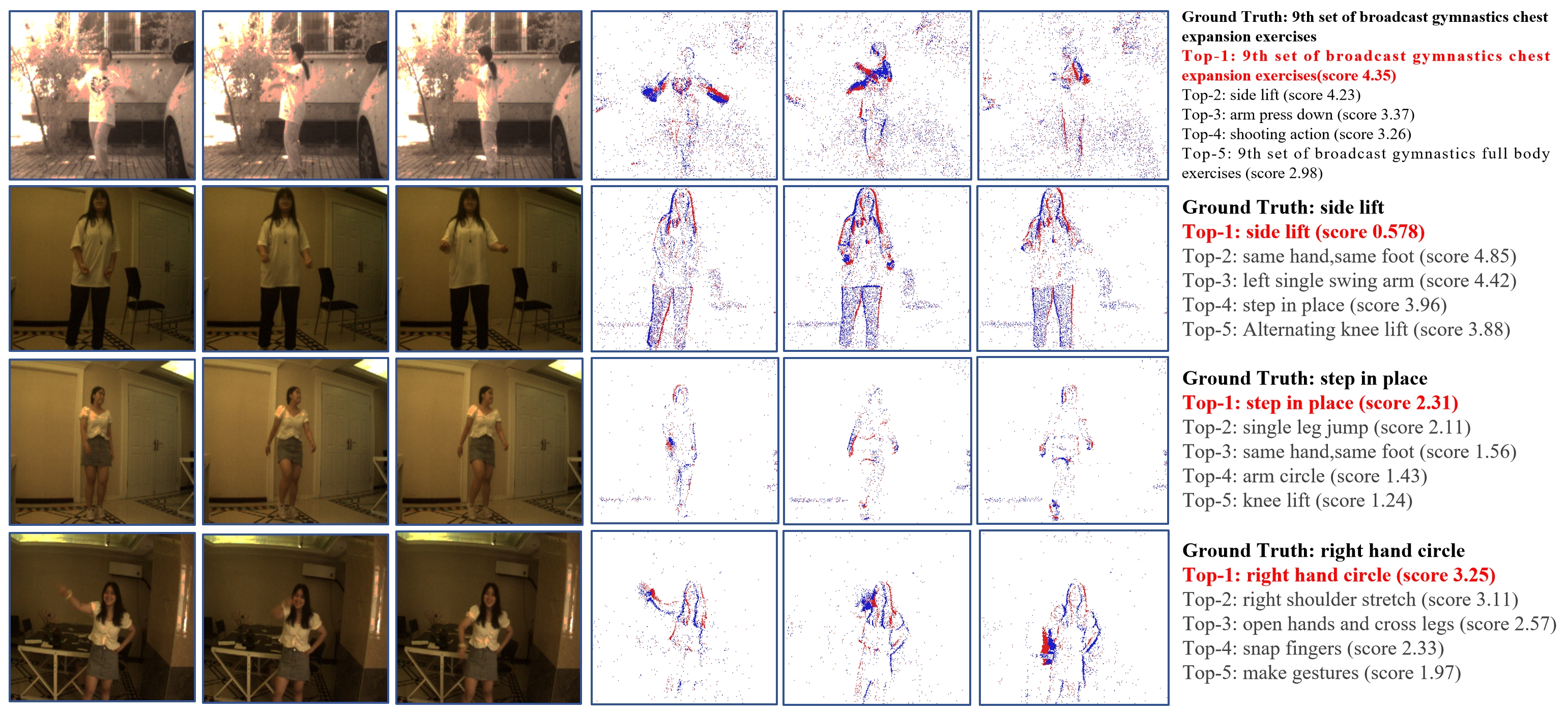}
\caption{Visualization of the top-5 predicted results on the HARDVS dataset.} 
\label{fig:top5_HARDVS_VIS}
\end{figure*}

\noindent $\bullet$ \textbf{Comparison with Variations of LoRA.~} 
In this sub-section, we explored different variants of LoRA, as outlined in Table~\ref{LoRA variant}. Interestingly, we discovered that training the model with the original LoRA configuration yielded the highest accuracy of 57.99\%. However, the rank of the original LoRA is immutable once set, and cannot be altered during training. DyLoRA~\cite{valipour2022dylora}, which trains LoRA blocks across a spectrum of ranks rather than a single fixed rank, resulted in a slightly lower accuracy of 57.35\%. 
Next, we experimented with a modification where we froze the $\textbf{A}$ matrix in LoRA, termed LoRA-FA~\cite{zhang2023lora}. This variant achieved an accuracy of 56.78\%, leading us to speculate that the reduced number of trainable parameters might be responsible for the suboptimal performance. We then introduced a linear layer between matrices $\textbf{A}$ and $\textbf{B}$, named LoRA-FA+, which slightly improved the accuracy to 57.22\%, but the enhancement was still modest. With the concept of a mixture of Experts, we also investigate the use of multiple LoRA experts combined with a routing mechanism. Despite this, the results did not exceed the original LoRA's performance, achieving an accuracy of only 57.01\%.

\begin{table}[!htp]
\scriptsize 
\tiny       
\caption{Comparison of different LoRA-based tuning strategies, LoRA-FA+ enhances LoRA-FA by incorporating an extra linear layer between matrices A and B.}
\label{LoRA variant} 
\resizebox{\columnwidth}{!}{
\begin{tabular}{c|c|c|c} 		
\hline 
\textbf{Algorithm}  & \textbf{LoRA}~\cite{hu2021lora} &\textbf{DoRA}~\cite{liu2024dora} &\textbf{LoRA-FA}~\cite{zhang2023lora}    \\  
Results    &57.99   & 57.65   &56.78            \\ 
\hline 
\textbf{Algorithm}  &\textbf{DyLoRA}~\cite{valipour2022dylora} &\textbf{LoRA-FA+}   &\textbf{MOE-LoRA}~\cite{luo2024moelora}  \\ 
Results    &57.35    &57.22 &57.01 \\
\hline 
\end{tabular}} 
\end{table}

\subsection{Efficiency Analysis} 
To give a more comprehensive evaluation of fully fine-tuning and our VELoRA, in this part, we primarily assess the efficiency of our proposed method based on four criteria: \textit{the number of parameters}, \textit{training time} (average duration per training epoch), \textit{memory consumption}, and \textit{FLOPS}.

As previously stated, throughout the training of our model, the parameters of the pre-trained model remain fixed, with only the weights of LoRA being updated. This approach significantly reduces the number of training parameters, conserves memory, and accelerates the training process. As depicted in Table~\ref{CAResults}, our fine-tuning method, in contrast to full fine-tuning, slashes the training parameters from 1719MB to a mere 7.02MB and frees up 2200MB of runtime memory. Additionally, our proposed method offers a quicker runtime and lower FLOPS, enhancing overall efficiency.

\begin{table*}[!htp] 
\centering 
\caption{Comparison between full fine-tuning and our proposed tuning strategy.}   
\label{CAResults} 
\begin{tabular}{c|c|c|c|c} 		
\hline 
\textbf{Method}  & \textbf{Parameters (MB)} &\textbf{Training Time (s)} &\textbf{Memory Cost (MB)}    &\textbf{FLOPs (GB)}   \\
\hline 
Full-tuning &1719       &4536       &19574        &30.98    \\
Ours        &7.02       & 4240      &17520        &41.35    \\
\hline 
\end{tabular}
\end{table*}

\subsection{Visualization} 
In this section, we present a quantitative analysis to deepen the interpretability of our algorithm. The subsequent subsections will provide insights into the feature embedding, the top-5 recognition outcomes, the RGB-Event frame differences, and the RGB activation maps, respectively.

\noindent $\bullet$ \textbf{Visualization of Feature Distribution Maps.} 
In this section, we visualize the feature distribution maps for different LoRA variants, including LoRA-FA~\cite{zhang2023lora}, MOE-LoRA~\cite{luo2024moelora}, and our proposed VELoRA. Our research reveals that the performance of the VELoRA model exhibits a moderate upgrade over the baseline model. Moreover, when contrasted with LoRA-FA, a significant performance boost is evident. These visual representations highlight the effectiveness of our proposed module in managing both RGB frames and event streams.

\noindent $\bullet$ \textbf{Visualization of the Frame Differences.~} 
In this section, we present visualizations of five sets of RGB frame differences (left) and Event frame differences (right) from the HARDVS dataset. Both modalities are processed uniformly; using the RGB modality as a case in point, there are varying degrees of differences between consecutive frames. Frame difference information can be used to extract key frames from a video, highlight certain image features, and thereby differentiate between the foreground and background.
As depicted in Figure~\ref{fig:frameDiff_VIS}, we have applied grayscale processing to adjacent frames, which facilitates a clearer observation of the action differences between frames. Concurrently, we have enhanced each frame-difference image, making it evident that our frame-difference images capture the essential action information between consecutive frames.

\noindent $\bullet$ \textbf{Visualization of the Activation Maps.~} 
As illustrated in Figure~\ref{fig:activation_VIS}, we provide a visualization of the original RGB image alongside the corresponding activation map from the HARDVS dataset. We showcase activation maps for images with varying levels of brightness and clarity. It is clear from these visualizations that our model is capable of precisely pinpointing the primary areas of activity. This demonstrates the effectiveness and robustness of our method, which can recognize actions and scenes across diverse conditions, thereby highlighting its generalizability.

\noindent $\bullet$ \textbf{Visualization of Recognition Results.} 
As shown in Figure~\ref{fig:top5_HARDVS_VIS}, we present the visualization of four sets of RGB-event images and their top-5 results on HARDVS datasets. From these images, we can observe that RGB images are easily affected by motion blur and lighting conditions, while the event stream can capture dynamic details and filter out static background noise. The results highlight the effectiveness of our suggested approach.


\subsection{Limitation Analysis} 
Although our model has achieved commendable experimental results, there is still a significant gap in recognizing every action with precision. The outcomes from these two datasets suggest that improving results solely through deep neural network models remains a formidable challenge. One possible reason is that the static representation of event streams may limit the effectiveness of temporal information. In future work, we plan to design a dynamic and learnable representation of event streams to further enhance the efficacy of the input data. Moreover, leveraging large models to analyze and reason about pedestrian actions is another approach worth considering to enhance the interpretability of our model.

\section{Conclusion} \label{sec::conclusion}
In this paper, we introduce a novel fine-tuning approach that integrates modality-specific and shared modality LoRA tuning strategy, enabling the model to retain sensitivity to individual modalities while also extracting shared information across various modalities, thereby improving the model's performance on multimodal tasks. Our model accepts three types of inputs, namely RGB, event, and frame differences. 
In the modality-specific tuning stage, we utilize a pre-trained large vision model to encode these inputs. It is important to note that although the encoders for these three branches are identical, their parameters are not shared. We contend that the three modalities are distinct and should not share parameters. 
In the modality-shared tuning stage, aimed at capturing shared information across different modalities, we incorporate the fused modality information into the final block of the Transformer and refine the original output. Our proposed method has indeed outperformed other state-of-the-art methods in terms of accuracy. Furthermore, our approach can dramatically reduce the number of training parameters in the model, thereby significantly conserving space and time resources.


\small{ 
\bibliographystyle{IEEEtran}
\bibliography{reference}
}

\vspace{-0.5cm}
\begin{IEEEbiography}[{\includegraphics[width=1in,height=1.25in, clip]
{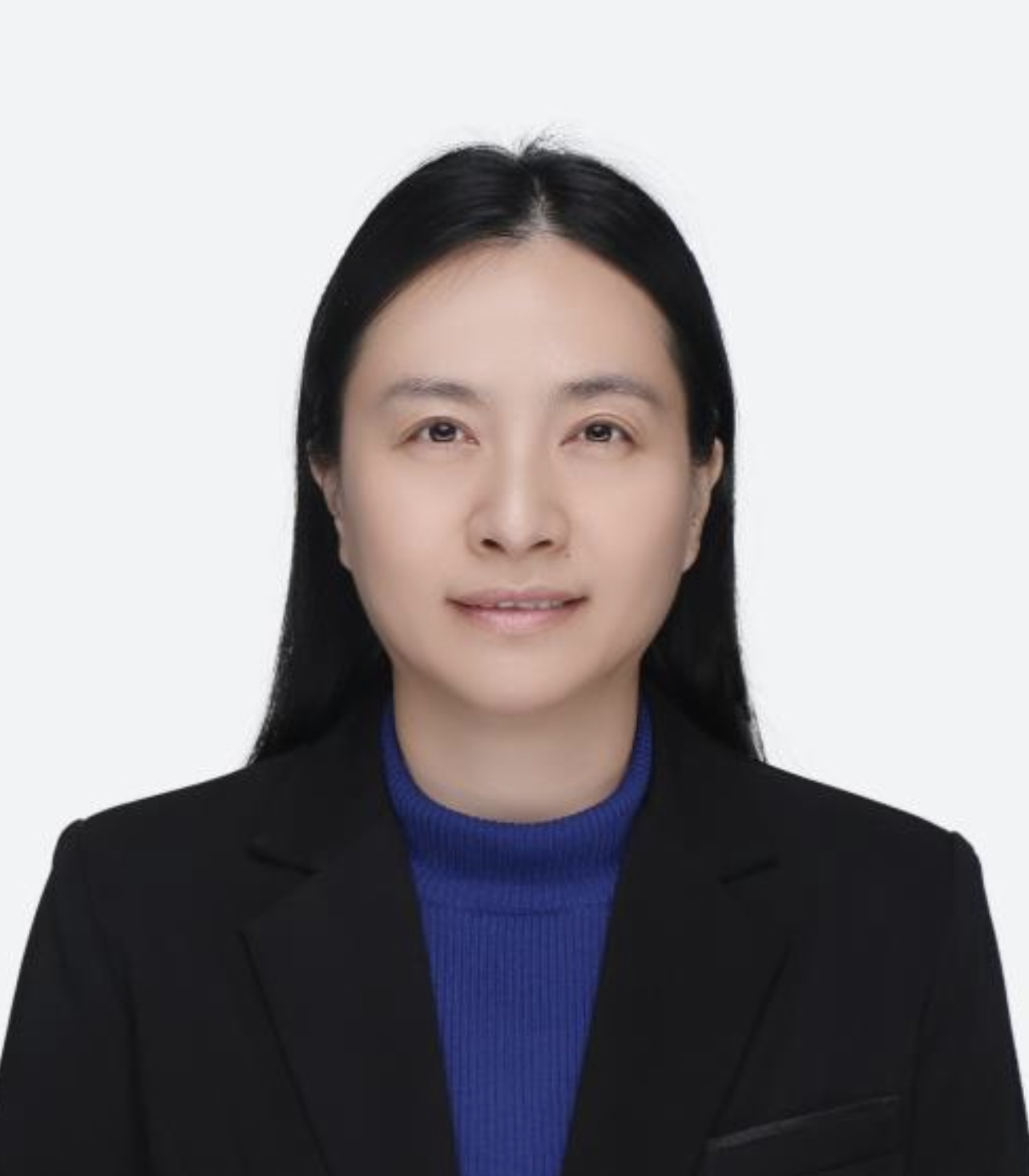}}]{Lan Chen} received the M.S. degree in Circuits and Systems from Anhui University, Hefei, China. She is currently a lecturer at the School of Electronic and Information Engineering, Anhui University, Hefei, China. Her research interests include computer vision, event-based vision, and deep learning. 
\end{IEEEbiography}

\vspace{-1cm}
\begin{IEEEbiography}[{\includegraphics[width=1in,height=1.25in, clip]{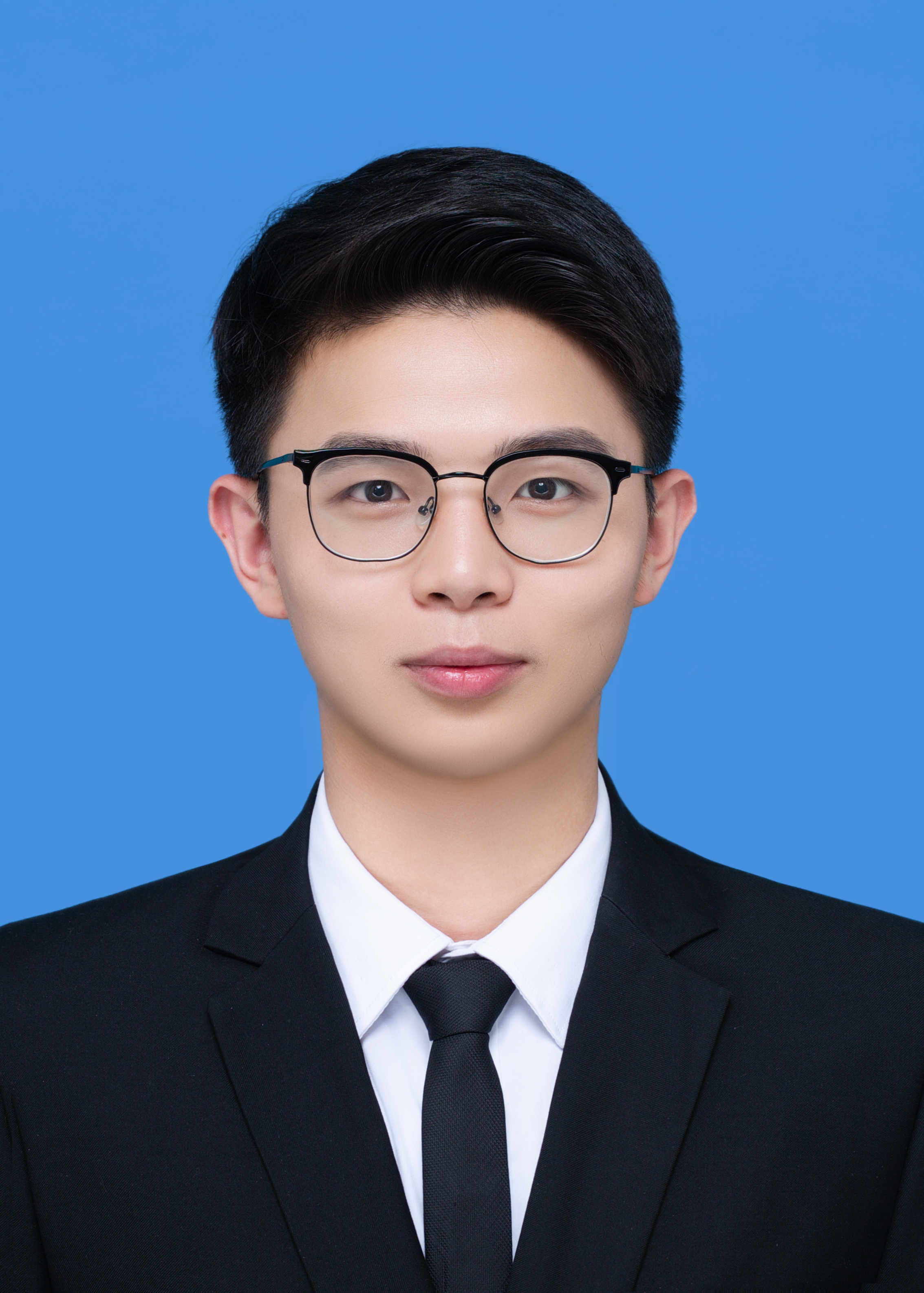}}]
{Haoxiang Yang} received B.E. degree in Computer Science and Technology from Hefei Normal University, Hefei, China. currently, He is pursuing a M.S. degree at the School of Computer Science and Technology, Anhui University,
Hefei, China. His current research interests include computer vision and deep
learning.
\end{IEEEbiography}

\vspace{-1cm}
\begin{IEEEbiography}[{\includegraphics[width=1.1in,height=1.25in, clip]{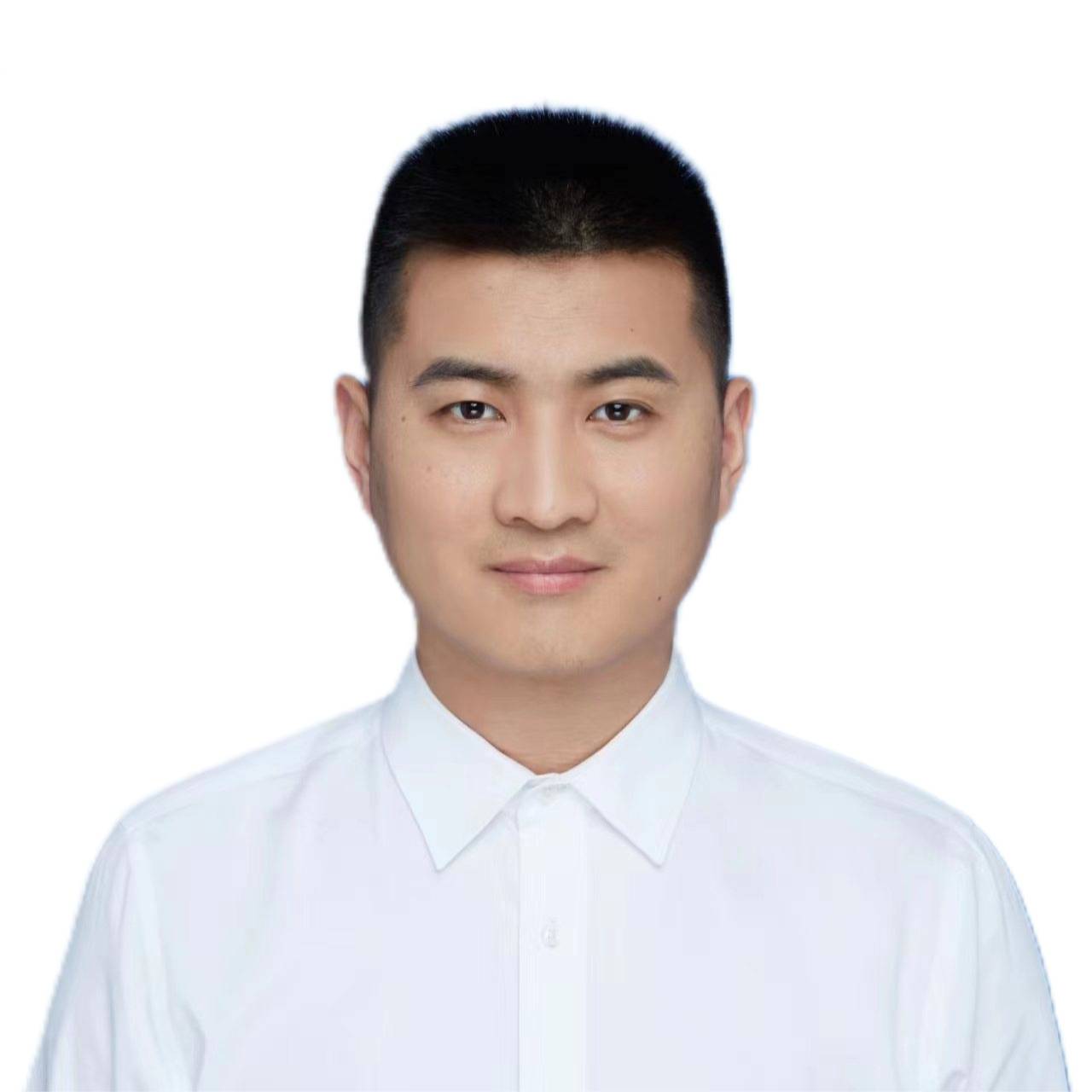}}]
{Pengpeng Shao} received Ph.D. degree from the Institute of Automation, Chinese Academy of Sciences, Beijing, China, in 2023. Currently, he is an assistant researcher in the Department of Automation, Tsinghua University, Beijing, China. His current research interests include natural language processing and temporal knowledge graphs.
\end{IEEEbiography}

\vspace{-1cm}
\begin{IEEEbiography}[{\includegraphics[width=1in,height=1.25in, clip]{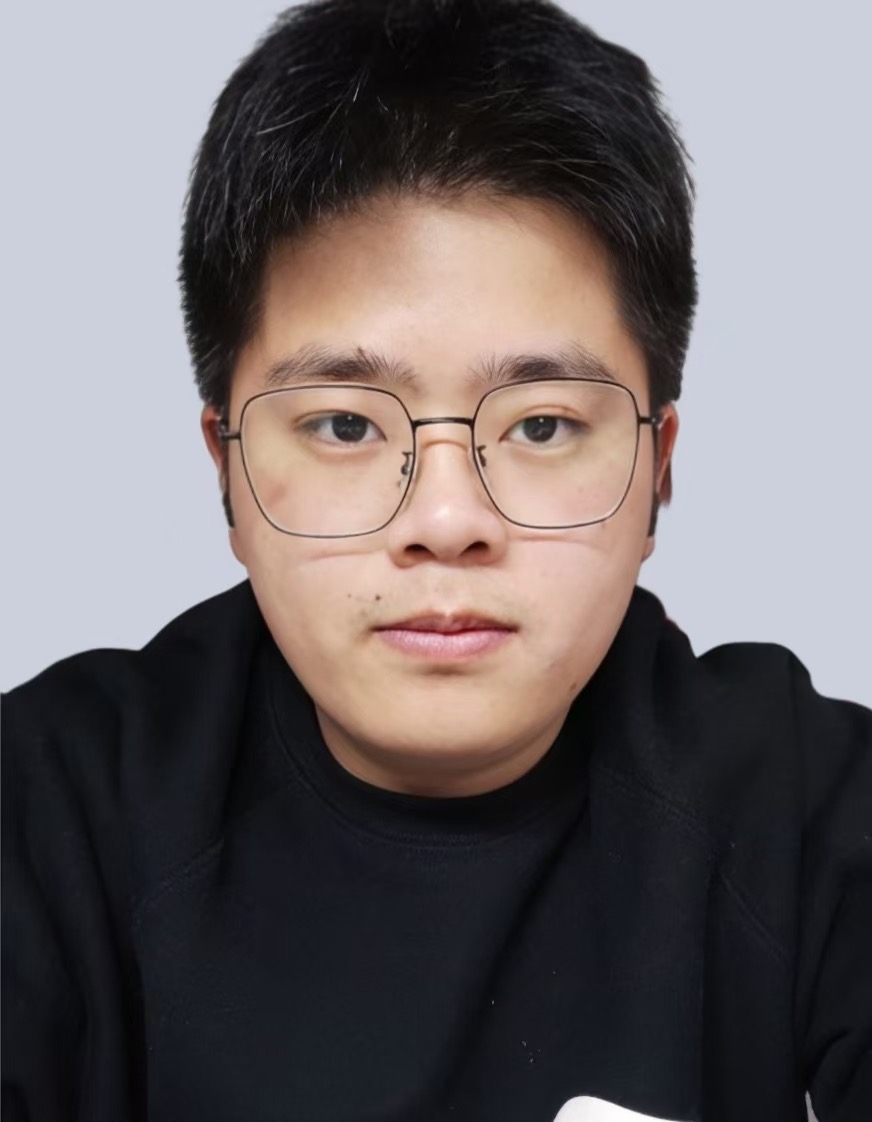}}]
{Haoyu Song} currently studies in the School of Computer Science and Technology at Anhui University, Hefei, China. He is a third-year undergraduate student and is expected to graduate in 2026. His current research interests include computer vision and deep learning.
\end{IEEEbiography}

\vspace{-1cm}
\begin{IEEEbiography}[{\includegraphics[width=1in,height=1.25in, clip]{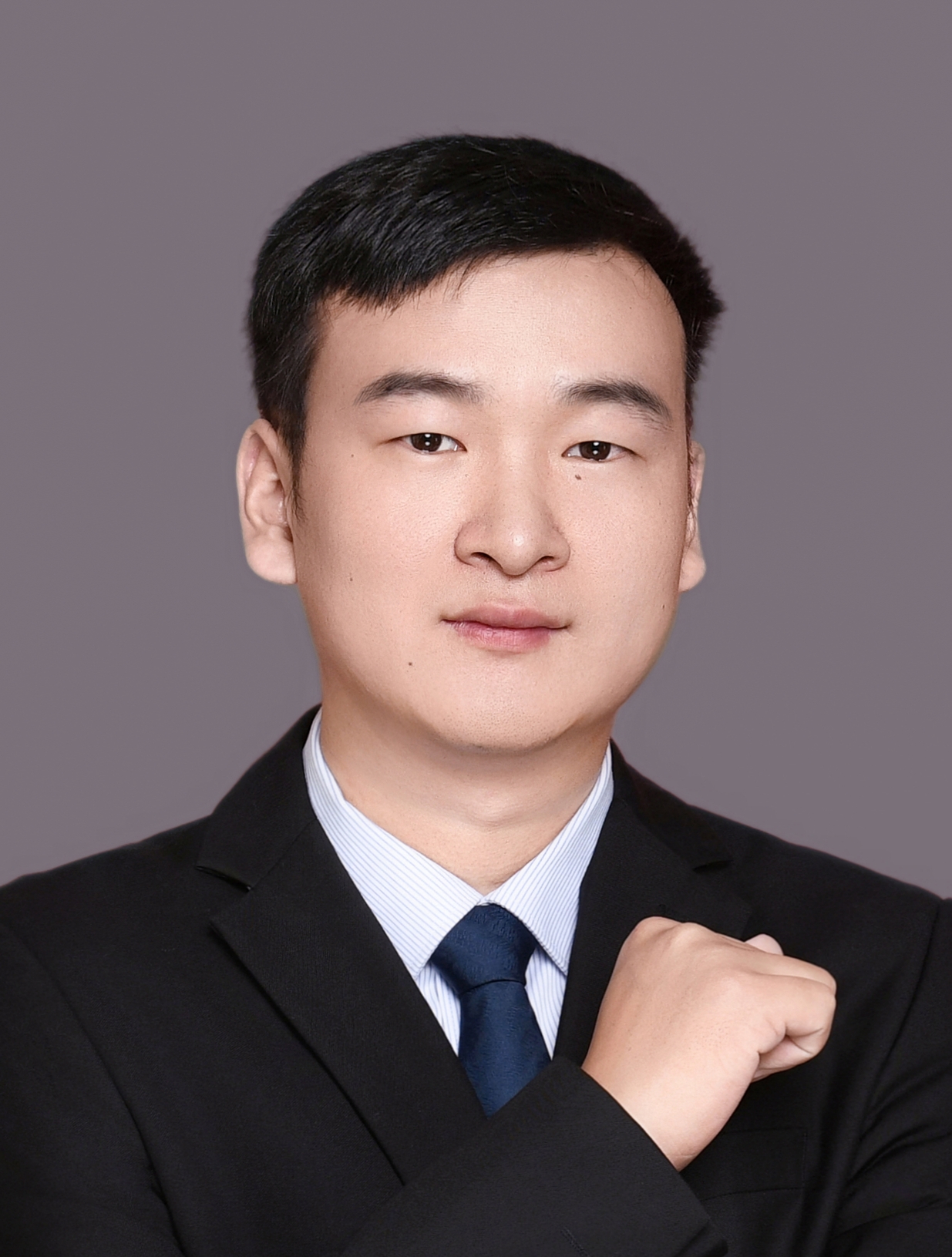}}]{Xiao Wang} (Member, IEEE) received the B.S.  degree from West Anhui University, Lu’an, China, in 2013, and the Ph.D. degree in computer science from Anhui University, Hefei, China, in 2019. From 2015 and 2016, he was a Visiting Student with the School of Data and Computer Science, Sun Yatsen University, Guangzhou, China. He has visited the UBTECH Sydney Artificial Intelligence Centre, Faculty of Engineering, The University of Sydney, Sydney, NSW, Australia, in 2019. He is a Post-Doctoral Researcher with the Peng Cheng Laboratory, Shenzhen, China, from 2020 to 2022. He is currently an Associate Professor with the School of Computer Science and Technology, Anhui University, Hefei, China. 
His current research interests include computer vision, event-based vision, and deep learning. 
Homepage: \url{https://wangxiao5791509.github.io/} 
\end{IEEEbiography}

\vspace{-1cm}
\begin{IEEEbiography} [{\includegraphics[width=1in,height=1.25in,clip,keepaspectratio]{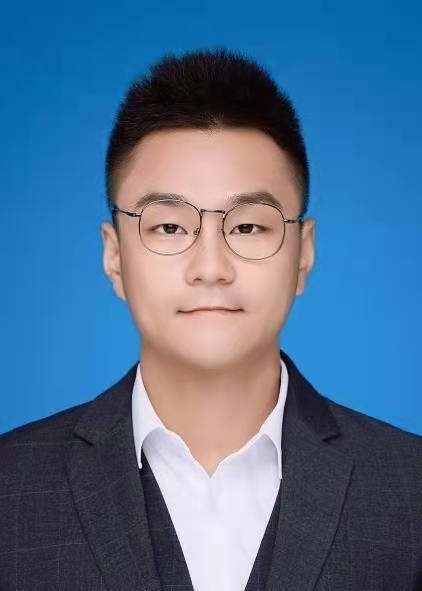}}]
{Zhicheng Zhao} received the B.E. degree from the National Pilot School of Software, Yunnan University, Kunming, China, in 2016, and the Ph.D. degree from the Computer Network Information Center, Chinese Academy of Sciences, Beijing, China, in 2021. He is currently a Lecturer with the Information Materials and Intelligent Sensing Laboratory of Anhui Province, Anhui Provincial Key Laboratory of Multimodal Cognitive Computation, School of Artificial Intelligence, Anhui University, Hefei, China. He also serves as a Postdoctoral Researcher with the 38th Research Institute of China Electronics Technology Group Corporation, Hefei. His research interests include computer vision and deep learning.
\end{IEEEbiography}

\vspace{-1cm}
\begin{IEEEbiography}[{\includegraphics[width=1in,height=1.25in,clip,keepaspectratio]{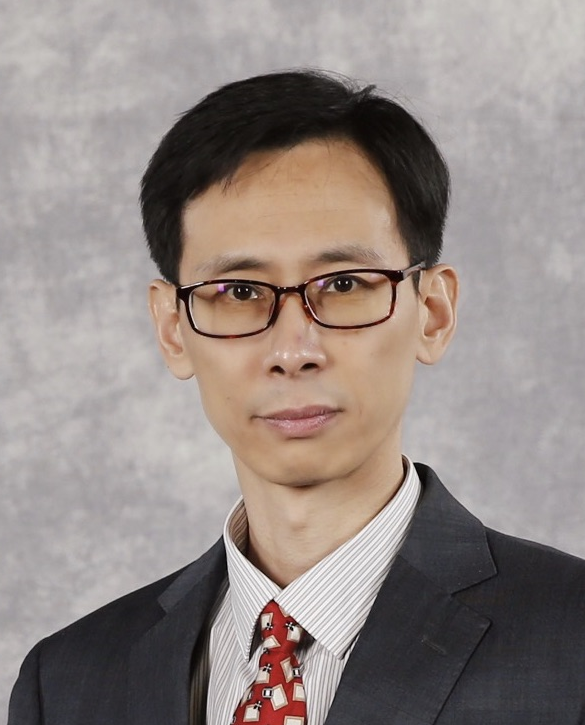}}] 
{Yaowei Wang} (Ph.D., Member, IEEE), is currently an associate professor with the Peng Cheng Laboratory, Shenzhen, China. He was a professor at National Engineering Laboratory for Video Technology Shenzhen (NELVT), Peking University Shenzhen Graduate School in 2019. From 2014 to 2015, he worked as an academic Visitor at the vision lab of Queen Mary University of London. He worked at the Department of Electronics Engineering, Beijing Institute of Technology from 2005 to 2019. He received his Ph.D. degree in Computer Science from the Graduate University of Chinese Academy of Sciences in 2005. His research interests include machine learning, multimedia content analysis and understanding. He is the author or coauthor of over 70 refereed journals and conference papers. He was the recipient of the second prize of the National Technology Invention in 2017 and the first prize of the CIE Technology Invention in 2015. His team was ranked as one of the best performers in the TRECVID CCD/SED tasks from 2009 to 2012 and in PETS 2012. He is a member of IEEE, CIE, CCF and CSIG.
\end{IEEEbiography}

\vspace{-1cm}
\begin{IEEEbiography}[{\includegraphics[width=1in,height=1.25in,clip,keepaspectratio]{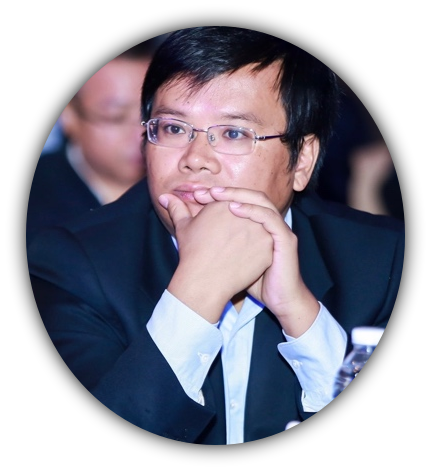}}] 
{Yonghong Tian} (Fellow, IEEE) is currently a Boya Distinguished Professor with the Department of Computer Science and Technology, Peking University, China, and is also the deputy director of Artificial Intelligence Research Center, PengCheng Laboratory, Shenzhen, China. His research interests include neuromorphic vision, brain-inspired computation and multimedia big data. He is the author or coauthor of over 200 technical articles in refereed journals such as IEEE TPAMI/TNNLS/TIP/TMM/TCSVT/TKDE/TPDS, ACM CSUR/TOIS/TOMM and conferences such as NeurIPS/CVPR/ICCV/AAAI/ACMMM/WWW. Prof. Tian was/is an Associate Editor of IEEE TCSVT (2018.1-), IEEE TMM (2014.8-2018.8), IEEE Multimedia Mag. (2018.1-), and IEEE Access (2017.1-). He co-initiated IEEE Int’l Conf. on Multimedia Big Data (BigMM) and served as the TPC Co-chair of BigMM 2015, and aslo served as the Technical Program Co-chair of IEEE ICME 2015, IEEE ISM 2015 and IEEE MIPR 2018/2019, and General Co-chair of IEEE MIPR 2020 and ICME2021. He is the steering member of IEEE ICME (2018-) and IEEE BigMM (2015-), and is a TPC Member of more than ten conferences such as CVPR, ICCV, ACM KDD, AAAI, ACM MM and ECCV. He was the recipient of the Chinese National Science Foundation for Distinguished Young Scholars in 2018, two National Science and Technology Awards and three ministerial-level awards in China, and obtained the 2015 EURASIP Best Paper Award for Journal on Image and Video Processing, and the best paper award of IEEE BigMM 2018. He is a senior member of IEEE, CIE and CCF, a member of ACM.
\end{IEEEbiography}

\end{document}